\documentclass[11pt]{article}

\usepackage{dialogue}
\usepackage{hyperref}

\usepackage{ifxetex}
\ifxetex
    \usepackage{fontspec}
    \setromanfont{Times New Roman}
\else
  \usepackage{cmap}
  \usepackage[T2A,T1]{fontenc}
  \usepackage[utf8]{inputenc}
  \usepackage{times}
  \usepackage{latexsym}
  \usepackage{substitutefont}
  \usepackage{pgf-pie}
  \usepackage{pgfplots}
  \pgfplotsset{compat=1.17}
  \substitutefont{T2A}{\familydefault}{cmr}
\fi

\usepackage[russian,british]{babel}
\usepackage{url}
\usepackage{pgf}

\usepackage{covington} 

\usepackage{hyperref}

\usepackage[frozencache,cachedir=.]{minted}

\newcommand{\russianexample}[2]{\foreignlanguage{russian}{\textit{#1}} (‘#2’)}
\newcommand{\corefmention}[2]{\{#1\}$_{#2}$}

\dialogfinalcopy 

\title{RuCoCo: a new Russian corpus with coreference annotation}

\author{Vladimir Dobrovolskii \\
  ABBYY \\
    \\
  {\tt v.dobrovolskii@abbyy.com} \\\And
  Mariia Michurina \\
  MIPT, RSUH \\
  Moscow, Russia \\
  {\tt marimitchurina@gmail.com} \\\AND
  Alexandra Ivoylova \\
  MIPT, RSUH \\
  Moscow, Russia \\
  {\tt a.m.ivoylova@gmail.com}}

\date{}

\begin{document}
\maketitle
\begin{abstract}
We present a new corpus with coreference annotation, Russian Coreference Corpus (RuCoCo). The goal of RuCoCo is to obtain a large number of annotated texts while maintaining high inter-annotator agreement. RuCoCo contains news texts in Russian, part of which were annotated from scratch, and for the rest the machine-generated annotations were refined by  human annotators. The size of our corpus is one million words and around 150,000 mentions. We make the corpus publicly available\footnote{\url{https://github.com/vdobrovolskii/rucoco}}.

\textbf{Keywords:} coreference corpus, coreference resolution, anaphora resolution, corpus annotation, Russian language

\end{abstract}

\selectlanguage{russian}
\begin{center}
  \russiantitle{RuCoCo: новый русскоязычный корпус кореференции}

  \medskip \setlength\tabcolsep{1cm}
  \begin{tabular}{cc}
    \textbf{Добровольский В.А.} & \textbf{Мичурина М.А.}\\
      ABBYY & МФТИ, РГГУ \\
        & Москва, Россия \\
      {\tt v.dobrovolskii@abbyy.com} & {\tt marimitchurina@gmail.com}\\
  \end{tabular}

  \begin{tabular}{c}
    \textbf{Ивойлова А.М.}\\
      МФТИ, РГГУ\\
      Москва, Россия\\
      {\tt a.m.ivoylova@gmail.com} \\
  \end{tabular}
  \medskip
\end{center}

\begin{abstract}
В этой статье мы представляем новый корпус кореференции для русского языка RuCoCo. Цель корпуса RuCoCo - получить большое количество размеченных текстов и одновременно с этим добиться высокого уровня согласия между аннотаторами. RuCoCo состоит из текстов новостей на русском языке, часть из которых была аннотирована с нуля, а для остальных текстов была выполнена машинная разметка и доработана аннотаторами-носителями языка. Размер нашего корпуса составляет один миллион слов и около 150 000 упоминаний. Корпус находится в открытом доступе.

  \textbf{Ключевые слова:} корпус кореференции, разрешение кореференции, разрешение анафоры, создание корпуса, русский язык
\end{abstract}
\selectlanguage{british}

\section{Introduction}
The task of coreference resolution was introduced at the Sixth Message Understanding Conference \cite{grishman1996message}, where the first dataset for coreference resolution task was introduced.
The dataset consisted of 25 articles from Wall Street Journal (30,000 words).
The annotation scheme was considered a standard until the release of ACE 2005 Multilingual Training Corpus for the 2005 Automatic Content Extraction (ACE) technology evaluation \cite{doddington-etal-2004-automatic}. The corpus included texts in English, Chinese and Arabic and contained around 650,000 words in total for the three languages. 

The MUC guidelines were domain-oriented, and their definition of a \emph{markable} (mention) was mostly syntactically motivated. But further developments in this area, starting with the ACE initiative, increasingly involved semantic factors, so that recent corpora with coreference annotation define markables based on semantic class restrictions.

Quite a lot of such corpora were created in the last two decades, their primary goals being to increase the size in order to satisfy the requirements of the data-driven approach and to improve inter-annotator agreement which in many cases is too low, especially when a dataset addresses more complex cases of coreference.

The most well-known corpus of this kind is OntoNotes 5.0 \cite{pradhan-etal-2013-towards}. OntoNotes contains texts of various genres in three languages: English, Arabic, and Chinese. The cumulative volume of this corpus is 2.9 million words (about 1.5 million being English). The average annotator agreement for OntoNotes is 91.8\% for normal coreference and 94.2\% for appositives \cite{hovy2006ontonotes}.

The authors of the ARRAU corpus \cite{poesio2008anaphoric,uryupina2020annotating} concentrate on "difficult" cases of anaphora: plural anaphora, abstract object anaphora, and ambiguous anaphoric expressions, so the corpus has bridging reference and discourse deixis annotated. It contains only English texts (although there is an Italian analogue LiveMemories \cite{rodriguez2010anaphoric}); its current size is 350,000 tokens. The inter-annotator agreement in ARRAU varies from 67\% (annotation of anaphoric ambiguities) to 95\% (annotation of complex anaphoric relations).

Thus, most of the largest corpora with coreference annotation contain predominantly English texts; however, with the growing interest in natural language processing of Non-English languages, corpora in other languages are being developed more often. As for the Russian language, there now exist two such datasets, one of them being RuCor \cite{toldova2014ru,toldova2015pre} and the other AnCor \cite{budnikov}.

RuCor contains texts from openly available sources, such as Russian OpenCorpora, Lib.ru and Lenta.ru (156,000 words in total). In this corpus the annotation process was conducted over morphosyntactically pre-processed texts. The annotation scheme differentiates between primary and secondary markables, according to Potsdam Coreference Scheme \cite{krasavina2007pocos}, where the primary markables are always annotated and represent specific references, while the secondary markables are annotated only if they are antecedents of any of the primary markables. Inter-annotator agreement for RuCor is 66\% (Cohen's Kappa) or 85\% (Mitkov's metric).

AnCor was created for the Ru-Eval competition in 2019 and contained 523 texts of various genres  from Russian OpenCorpora (193,000 words in total). Named entities, common NPs and pronouns were annotated; the inter-annotator agreement for this dataset is 62.7\% (75.5\% agreement of both annotators and the final version).

As can be seen, although there are plenty of different corpora with coreference annotation, the largest and the most complex ones do not contain texts in Russian, and as for the Russian corpora, they are significantly smaller than the English ones, besides, their inter-annotator agreement is lower.

Therefore our main goals were to create a sufficiently big Russian corpus which would contain annotation of at least some difficult cases of anaphora with the inter-annotator agreement being high enough compared to OntoNotes and ARRAU.

\section{RuCoCo: Russian Coreference Corpus}
\subsection{Data}
We utilize the news stories published by NEWSru.com\footnote{https://www.newsru.com/} as our source of text data. The texts were automatically collected and processed in the following way:
\begin{enumerate}
    \item Any texts containing videos or embedded widgets from other websites were discarded as well as any texts marked as promotions.
    \item Then the texts were converted to plain text format and cleared of any remaining HTML artifacts.
    \item Texts that contained fewer than 20 tokens were also discarded, because they mostly consisted of a heading and a follow-up link only.
    \item We then uniformly sampled one million words worth of texts across all text lengths and news categories. The total number of sampled texts is 3075.
\end{enumerate}

\begin{figure}[h] 
    \centering
    \includegraphics[width=0.8\linewidth]{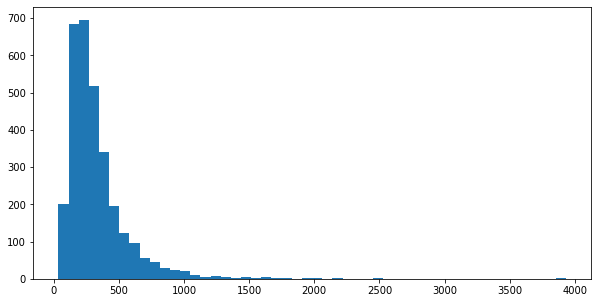}
    \caption{Distribution of text lengths in the sampled data.}
\end{figure}

\begin{figure}[h] 
    \centering
    \includegraphics[width=0.8\linewidth]{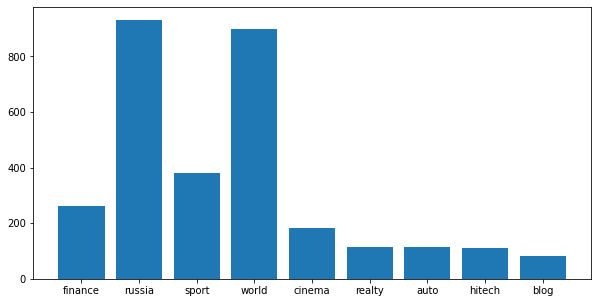}
    \caption{Distribution of news categories in the sampled data.}
\end{figure}

\subsection{Annotation layer}
The first release of RuCoCo covers identity (and in some cases, near-identity) coreference of noun phrases and pronouns. We do not annotate singletons, which means that each mention is linked to at least one other mention. We do not assign any attributes to the markables.

{\bf Mentions:} We treat all noun phrases as potential mentions. Additionally the following types of pronouns are annotated:
\begin{itemize}
    \item personal, possessive and reflexive pronouns;
    \item reciprocal pronouns, such as \russianexample{друг друга}{each other};
    \item relative pronouns;
    \item interrogative pronouns.
\end{itemize}

However, at this point we do not annotate coreference links with adjectives, clauses and expressions of time, all of which are going to be treated as valid mentions in the second revision of the corpus.

{\bf Mention boundaries:}
In most cases full noun phrases are annotated. To avoid overlapping of mentions referring to the same entity, participle and relative clauses that depend on the mention head are not included in mention boundaries. Therefore, in the following example there is no overlapping: \russianexample{\corefmention{клиент}{0}, \corefmention{который}{0} хотел пополнить \corefmention{свой}{0} счет}{\corefmention{a customer}{0}, \corefmention{who}{0} wanted to top up \corefmention{their}{0} account}. Parenthesis is not annotated unless it contains an independent clause, in which case it is treated as a regular sentence.

{\bf Coreference and anaphora: }
Coreference is annotated only for mentions of concrete entities. For generic mentions and mentions of abstract entities, events and properties we only annotate anaphora:
\russianexample{Может ли машина действовать разумно? Может ли \corefmention{машина}{0} обладать сознанием? Может ли \corefmention{она}{0} чувствовать?}{Can a machine act intelligently? Can \corefmention{a machine}{0} have a consciousness? Can \corefmention{it}{0} feel how things are?}. Here, the first mention of \russianexample{машина}{a machine} is not annotated as coreferent with other mentions, because it is a generic mention.

{\bf Ellipsis: } Mentions with elided heads are not annotated, as it would create ambiguity: \russianexample{Это твоя сестра или \underline{\corefmention{Даниэля}{0}}? Это \corefmention{сестра \corefmention{Даниэля}{0}}{1}, \corefmention{она}{1} приехала на выходные.}{Is this your sister or \corefmention{Daniel's}{0}? This is \corefmention{\corefmention{Daniel's}{0} sister}{1}, \corefmention{she}{1} came for the weekend.}. In the example above, the underlined mention could be recovered as \russianexample{сестра Даниэля}{Daniels' sister}, but we do not annotate it as referring to entity \#1, because there would be two identical mentions referring to different entities.

{\bf Split antecedents: } In RuCoCo, we annotate split antecedents as a means of dealing with the following challenges:
\begin{itemize}
    \item Mentions referring to multiple referents: \russianexample{\corefmention{Премьер-министр}{0} и \corefmention{госпожа Саймондс}{1} поженились вчера днем, небольшая церемония прошла в Вестминстерском соборе. \corefmention{Пара}{0,1} отпразднует свадьбу с семьей и друзьями следующим летом.\footnote{https://www.newsru.com/world/30may2021/bjohnson.html}}{\corefmention{Prime Minister}{0} has married \corefmention{Carrie Symonds}{1} yesterday afternoon in a "small ceremony" at Westminster Cathedral. \corefmention{The couple}{0,1} would celebrate again with family and friends next summer.}.
    \item Coordinate dependents: \russianexample{\corefmention{Сборные России и Канады}{0,1} ранее ни разу не встречались в финалах чемпионатов мира. <..> \corefmention{Отечественные хоккеисты}{0} победили \corefmention{канадцев}{1} со счетом 5-3 в Стокгольме в 1989 году.}{\corefmention{National teams of Russia and Canada}{0,1} have not played in IIHF finals before. <..> \corefmention{The Russian team}{0} defeated \corefmention{the Canadians}{1} 5-3 in Stockholm in 1989.}
\end{itemize}
Further in the text we refer to mentions linked to split antecedents as \emph{plural anaphors} and to entities built from such mentions as \emph{plural anaphor entities}. The number of such entities in the corpus can be seen in Table \ref{total-stats}.

\begin{table}[h]
\begin{center}
\begin{tabular}{|l|rr|rrr|}
\hline \bf Category & \bf Words & \bf Mentions & \bf Entities & \bf PA-Entities & \bf APA-Entities \\ \hline

russia  &   352,672     &   55,338  &   13,891  &   1,083   (7.8\%) &   2,471   (17.8\%)    \\
world   &   311,445     &   50,283  &   12,660  &   1,045   (8.3\%) &   2,122   (16.8\%)    \\
finance &   94,015      &   11,739  &   3,020   &   176     (5.8\%) &   447     (14.8\%)    \\
sport   &   80,352      &   11,807  &   3,331   &   279     (8.4\%) &   705     (21.2\%)    \\
cinema  &   53,645      &   8,003   &   2,116   &   167     (7.9\%) &   431     (20.4\%)    \\
realty  &   34,227      &   4,509   &   1,274   &   72      (5.7\%) &   184     (14.4\%)    \\
hitech  &   31,365      &   3,895   &   1,080   &   77      (7.1\%) &   150     (13.9\%)    \\
auto	&   24,735      &   2,914   &   881     &   40      (4.5\%) &   94      (10.7\%)    \\
blog    &   17,649      &   1,917   &   624     &   39      (6.3\%) &   94      (15.0\%)    \\

\hline
Total   &   1,000,105   &   150,405 &   38,877  &   2,978 (7.7\%)   &   6,698 (17.2\%)  \\

\hline
\end{tabular}
\end{center}
\caption{Number of words, extracted mentions, entities, plural-anaphor (PA) entities and antecedent-of-plural-anaphor (APA) entitities across the news categories in RuCoCo.}
\label{total-stats}
\end{table}

{\bf Metonymy: } Linking of metonymies is allowed: \russianexample{\corefmention{Лондон}{0} и \corefmention{Брюссель}{1} официально объявили о соглашении по Brexit. \corefmention{Евросоюзу}{1} и \corefmention{Великобритании}{0} удалось выработать соглашение об отношениях после Brexit.}{\corefmention{London}{0} and \corefmention{Brussels}{1} have announced a Brexit trade deal. \corefmention{The European Union}{1} and \corefmention{the United Kingdom}{0} have agreed on a post-Brexit trade deal.}.

{\bf Corpus format: } RuCoCo is distributed as a collection of JSON-formatted files. An entity is represented as a list of character offset pairs. Antecedents of plural anaphor entities are listed in the "includes" section.

\begin{listing}[h]
\begin{minted}[frame=single,
               framesep=3mm,
               breaklines,
               tabsize=4]{js}
{
    "entities" : [[[31, 34]], [[39, 42], [100, 103]], [[71, 75]]],
    "includes" : [[], [], [0, 1]],
    "text": "At half-past nine, that night, Tom and Sid were sent to bed, as usual. They said their prayers, and Sid was soon asleep.\n"
}
\end{minted}
\caption{JSON-formatted annotation of the following example: \textit{At half-past nine, that night, \corefmention{Tom}{0} and \corefmention{Sid}{1} were sent to bed, as usual. \corefmention{They}{0,1} said their prayers, and \corefmention{Sid}{1} was soon asleep.}}
\label{json-example}
\end{listing}

\section{Corpus annotation}

\subsection{Metrics}
There exist a number of coreference evaluation metrics, such as \emph{MUC} \cite{vilain-etal-1995-model}, \emph{B$^3$} \cite{bagga-baldwin-1998-entity-based}, \emph{CEAF} \cite{luo-2005-coreference}, \emph{BLANC} \cite{blanc-2011} and others. Since the CoNLL-2012 shared task \cite{pradhan-etal-2012-conll}, the average score of \emph{MUC}, \emph{B$^3$} and \emph{CEAF$_e$}, has become a de-facto standard way to evaluate coreference resolution systems. However, several shortcomings of these three metrics were demonstrated by \newcite{moosavi-strube-2016-coreference}, who also introduced \emph{LEA}, a coreference evaluation metric designed to overcome those shortcomings. \emph{LEA} of a set of entities $K$ is computed as:
\begin{equation}
    \frac{
        \sum_{e_i \in E} (importance(e_i) \times resolutionScore(e_i))
    }{
        \sum_{e_j \in E} importance(e_j)
    }
\end{equation}
where $importance(e) = |e|$ and the resolution score of entity $k_i$ is calculated against the response set of entities $R$ as follows:
\begin{equation}
    resolutionScore(k) = \sum_{r_j \in R} \frac{ link(k_i \cap r_j) }{ link(k_i) }
\end{equation}
Here, $link(e)$ calculates the number of unique coreference link within $e$: $link(e) = |e| \times (|e| - 1) / 2$.

We adopt \emph{LEA} as our primary metric for measuring inter-annotator agreement and evaluating the neural coreference resolution model. As \emph{LEA} does not support split antecedents out of the box, we modify the metric in the following way: for each plural anaphor entity we additionally calculate the scores of a special dummy entity with $importance$ set to be the number of antecedent entities and $resolutionScore$ computed based on the directed links between the plural anaphor entity and its antecedent entities.

The corpus was annotated by a team of 20 students of General Linguistics. The annotators were chosen based on trials that involved annotating documents of up to 1500 words. Each of the resulting documents was compared to the gold annotation using the \emph{LEA} metric. The passing score was set to 0.9; the passing rate was 67\%. Five of the annotators with the highest annotation quality were later appointed as moderators.

\subsection{Neural pre-annotator}
To speed up the annotation process, we developed a neural coreference resolution model to pre-annotate the texts. The model is based on the architecture proposed by \newcite{lee-etal-2018-higher} and improved by \newcite{joshi-etal-2019-bert} with the following differences:
\begin{itemize}
    \item We use the Russian version of RoBERTa \cite{Liu2019RoBERTaAR} pretrained by Sber AI\footnote{https://github.com/sberbank-ai/model-zoo}.
    \item We replace the neural mention extraction module with a rule-based syntactic mention extractor built on top of spaCy \cite{spacy2}. This allows us to explicitly define what a mention is instead of relying on neural networks for mention extraction.
    \item Following \newcite{dobrovolskii-2021-word}, we represent mentions using only weighted sums of the subtoken embeddings that constitute the mention.
\end{itemize}
To train the model, we used the automatically merged annotations obtained during the early phases of annotation. We ignored plural anaphors and used the original \emph{LEA} to evaluate the pre-annotation quality. The model performed at 0.62 F1 after being trained on 100,000 words, at 0.68 F1 after being trained on 400,000 words and at 0.73 F1 after training on the whole dataset of 1,000,000 words.

\subsection{Annotation process}
The annotation process consisted of two steps: the first 100,000 words were annotated from scratch, i.e. the task was to identify and link all coreferent mentions in raw texts; the remaining 900,000 words were first pre-annotated by a neural coreference resolution model and the annotators were asked to correct the resulting documents.

Each text in the corpus was annotated by two annotators and then finalized by a moderator who received an automatic merge of the two versions with differences highlighted.
Additionally, 3500 words of each annotator were manually checked by the authors of the markup scheme to provide feedback on an early stage.

\subsection{Inter-annotator agreement}
We measured the inter-annotator agreement and found it to be 0.759 F1. Because the annotators do not have a closed set of mentions to link, we suspect that some of the differences between annotations can be attributed to lack of attention. To eliminate this factor, we conducted the following experiment on a subset of the data approximately 50,000 words in size: each annotator was given back their own annotations automatically merged with the other annotation versions. The annotators were asked to independently correct the documents. The resulting inter-annotator agreement was 0.890 F1.

\subsection{Disagreement analysis}

We analysed discrepancies of the two phases of corpus annotation: 1) from scratch (50 random texts, about 16,000 words examined) and 2) pre-tagged annotation (158 random texts, 50,000 words examined). Discrepancies were divided into several categories:
\begin{itemize}
    \item missing/redundant coreference cluster;
    \item missing/redundant markable;
    \item missing/redundant anaphoric chain;
    \item plural anaphors with split antecedents;
    \item mentions referred to different entities;
    \item NP borders.
\end{itemize}

\begin{figure}[h]
\begin{tikzpicture}
	\begin{axis}[
	xbar stacked,
	symbolic y coords={Pre-tagged annotation,
    Neural model annotation,
    Annotation from scratch},
    legend style={at={(0.5,-0.20)},
      anchor=north,legend columns=3,
      column sep=5pt},
    legend cell align={left},
    xmin=0,
    xmax=100,
	ytick=data,
    height=150,
    width=350,
    bar width=19pt,
    ]
	\addplot coordinates
		{(39.11,Annotation from scratch) (31.88,Neural model annotation) (38.96,Pre-tagged annotation)};
	\addplot coordinates
		{(20.66,Annotation from scratch) (35.35,Neural model annotation) (28.6,Pre-tagged annotation)};
	\addplot coordinates
		{(3.33,Annotation from scratch) (13.86,Neural model annotation) (8.6,Pre-tagged annotation)};
	\addplot coordinates
		{(8.89,Annotation from scratch) (7.79,Neural model annotation) (8.39,Pre-tagged annotation)};
	\addplot coordinates
		{(6.44,Annotation from scratch) (3.98,Neural model annotation) (7.87,Pre-tagged annotation)};
	\addplot coordinates
		{(21.55,Annotation from scratch) (7.1,Neural model annotation) (7.56,Pre-tagged annotation)};

	\legend{\strut Coreference cluster, \strut Markable, \strut Anaphoric chain, \strut Plural anaphors, \strut Different entities, \strut NP borders}
	\end{axis}

\end{tikzpicture}
\caption{Comparison of discrepancies on all annotation steps, \%}
\label{wholeDisargeement}

\end{figure}
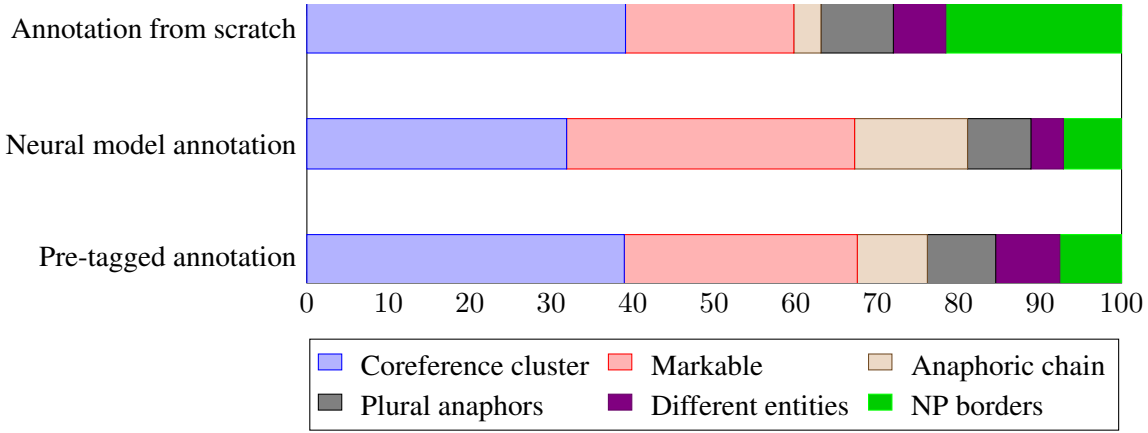

To make the comparison more informative, we carried out the error analysis of the neural model used for pre-tagging, although we need to keep in mind that after the first 100,000 words were checked, we made a number of minor clarifications and changes in the guidelines to facilitate the work of our annotation team. See the comparison of discrepancies in annotation from scratch, model errors and pre-tagged texts in Figure~\ref{wholeDisargeement}.

By \textbf{missing/redundant coreference cluster} we mean all cases when one of the two annotators skipped the whole cluster or marked up an unnecessary entity. It is the most frequent type when annotators disagree (about 39\% for both annotation stages). There was no closed set of entities, moreover, for abstract and generic entities, events or referents denoting open sets (so-called non-concrete entities) only anaphora must be annotated. Thus, annotators should decide whether the entity is concrete or non-concrete. They disagree on the following examples: locations without proper names like \foreignlanguage{russian}{\textit{кризисный регион}} (‘crisis area’), \foreignlanguage{russian}{\textit{жилой квартал}} (‘residential area’), and also when locations are nested in an organisation name: \foreignlanguage{russian}{\textit{Россия}} (‘Russia’) in \foreignlanguage{russian}{\textit{Министр транспорта России}} (‘Minister of Transport of Russia’) or in \foreignlanguage{russian}{\textit{Верховном суде РФ} (‘Supreme Court of the Russian Federation’)}. Some other popular types are events like \foreignlanguage{russian}{\textit{концерт в Москве}} (‘the concert in Moscow’), \foreignlanguage{russian}{\textit{чемпионат России по хоккею}} (‘Russian ice hockey championship’) and some abstract entities that are very similar to events as they have participants like \foreignlanguage{russian}{\textit{контракт}} (‘contract’), \foreignlanguage{russian}{\textit{уголовное дело}} (‘criminal case’).

\textbf{Missing/redundant markable} (about 20\% and 28\% respectively) is the case when an annotator missed one or several mentions, although the coreference cluster is there in both annotation versions. For these cases we examined types of NPs missed by one annotator in the annotation from scratch stage, having preserved the taxonomy as in \cite{toldova2015pre} in order to compare them. See Table~\ref{tab:NPTypes} to check numbers. We can observe that both annotation groups of students tend to miss noun groups (i.e. noun phrases headed by a noun) more than any other NP type.

\begin{table}[h]
	\begin{center}
	\begin{tabular}{r r r}
	\hline
	NP Type & Our Data, \% & Toldova et al., 2015, \%\\
	\hline
	Reflexive pronouns	&    4.73    & 3.76 \\
	Relative pronouns	&    1.77	& 6.20 \\
	Anaphoric pronouns	&   4.73 	& 12.47 \\
	Possessive pronouns	&   2.37 	& 6.48 \\
	Noun groups  &    85.2	& 71.08 \\
	Adverbs (here/there) & 1.18 & 0.00 \\
	\hline
	\end{tabular}
	\end{center}
	\caption{Types of missed NPs}
	\label{tab:NPTypes}
\end{table}

As for \textbf{missing/redundant anaphoric chain} (3.3\% and 8.6\%) i.e. chains with abstract or generic entities where only anaphora resolution was performed, annotators mostly missed chains containing a relative pronoun \foreignlanguage{russian}{\textit{который}} (‘which/that’) as an anaphoric element e.g. \foreignlanguage{russian}{\textit{срок, до которого}} (‘the deadline by which’), \foreignlanguage{russian}{\textit{той политической линии, которую}} (‘the policy that’).

In \textbf{plural anaphors with split antecedents} (9\% out of all discrepancies, both stages), the most common discrepancy is a missing relation between a person and a group of people: a son and a family, \foreignlanguage{russian}{\textit{Кондолиза Райс}} (‘Condoleezza Rice’), \foreignlanguage{russian}{\textit{сенатор Хиллари Клинтон}} (‘Senator Hillary Clinton’) and \foreignlanguage{russian}{\textit{политики}} (‘politicians’). Less frequent cases of disagreement are the following: part-whole relations (which are not annotated as split anaphora) and entities denoting several items with part of these items as split antecedents: \foreignlanguage{russian}{\textit{50 терактов}} (‘50 terror attacks’) and \foreignlanguage{russian}{\textit{20 терактов}} (‘20 terror attacks’).

\textbf{Mentions referred to different entities} (6.4\% and 7.9\%) include cases where one or several mentions were assigned to different clusters by annotators in some confusing contexts (e.g. pronouns) or one annotator labelled some mentions in one and the same chain while the other one has divided it into several chains e.g. cases with metonymy like \foreignlanguage{russian}{\textit{Пхеньян}} (‘Pyongyang’) and \foreignlanguage{russian}{\textit{КНДР}} (‘North Korea’), \foreignlanguage{russian}{\textit{Израиль}} (‘Israel’) and \foreignlanguage{russian}{\textit{Израильская армия}} (‘Israel Defense Forces’).

Disagreement on \textbf{NP borders} covers 21\% of discrepancies in the first stage and substantially less on the pre-tagged stage (7.5 \%). We may assume that it may be due to the ability of our model to find correct borders or that it is due to the clarified guidelines of syntactic ambiguities we made before the second annotation stage: we have highlighted that in all such cases the maximum NP border must be annotated. This category presupposes cases where annotators excluded modifiers as in \foreignlanguage{russian}{\textit{изменения}} (‘changes’) vs. \foreignlanguage{russian}{\textit{самые существенные изменения}} (‘the most significant changes’), complements e.g. \foreignlanguage{russian}{\textit{Банк}} (‘the Bank’) vs. \foreignlanguage{russian}{\textit{Банк России}} (‘the Bank of Russia’) and less often appositives: \foreignlanguage{russian}{\textit{Берт Ньюборн}} (‘Burt Neuborne’) vs. \foreignlanguage{russian}{\textit{Берт Ньюборн, профессор права Университета Нью-Йорка}} (‘Burt Neuborne Professor of Civil Liberties at New York University’).

This analysis was presented to the moderators so that they would know what to pay attention to. Despite all these discrepancies, the resulting inter-annotator agreement is still 0.890 F1 and all the disagreements were resolved by our moderators.

\section{Conclusion}
The result of our work is the Russian Coreference Corpus, which is the largest corpus with coreference annotation for Russian so far. We managed to achieve almost 90\% inter-annotator agreement; we also analyzed the most common disagreements between our annotators so that we know what issues are to be solved.
Further developments will include annotating more difficult cases of anaphora as well as increasing the size and genre diversity of the corpus.

\section*{Acknowledgements}
We are grateful to our annotation team from General Linguistics Department of RSUH for their hard work, attentive approach to the project and immense help in discussions. We would also like to thank Prof. Svetlana Toldova and Evgeniya Inshakova for their useful observations and helpful advice.

\bibliography{dialogue.bib}
\bibliographystyle{dialogue}

\end{document}